\documentclass[letterpaper, 10 pt, conference]{ieeeconf}  
\IEEEoverridecommandlockouts
\usepackage{graphics} % for pdf, bitmapped graphics files
\usepackage{graphicx}
\usepackage{epsfig} % for postscript graphics files
\usepackage{mathptmx} % assumes new font selection scheme installed
\usepackage{times} % assumes new font selection scheme installed
\usepackage{amsmath} % assumes amsmath package installed
\usepackage{amssymb}  % assumes amsmath package installed
\usepackage{multirow} %multirow for format of table
\usepackage[ruled,linesnumbered]{algorithm2e}

\usepackage[normalem]{ulem} %to strike the words

\usepackage[acronym]{glossaries}
\newacronym[
    shortplural={GPUs},
    longplural={Graphics Processing Units}
] {gpu}{GPU}{Graphics Processing Unit}

\newacronym[] {cnn}{CNN}{Convolutional Neural Network}

\newacronym[] {rc}{RC}{Radio Control}

\newacronym[] {sota}{SOTA}{state-of-the-art}

\newacronym[] {lr}{LR}{learning rate}

\newacronym[] {ohem}{OHEM}{Online Hard Example Mining}

\newacronym[] {pvtl}{PVTL}{Pedestrian and Vehicle Traffic Lights}

\newacronym[] {pov}{POV}{point-of-view}

\newacronym[] {bvip}{BVIP}{blind and visually impaired people}

\newacronym[] {hog}{HOG}{histogram of oriented gradients}

\newacronym[] {svm}{SVM}{support vector machine}

\newacronym[] {fps}{FPS}{frame per second}

\newacronym[
    shortplural={IoUs},
    longplural={Intersection over Unions}
] {iou}{IoU}{Intersection over Union}

\newacronym[] {miou}{mIoU}{mean Intersection over Union}

\newacronym[
    shortplural={MLPs},
    longplural={Multilayer Perceptrons}
] {mlp}{MLP}{Multilayer Perceptron}

\usepackage{caption}
\usepackage{subcaption}
\let\labelindent\relax
\usepackage[inline]{enumitem}
\usepackage{float}

\usepackage{xcolor}
\usepackage{paralist}

\usepackage{tabularx}
\usepackage{array}
\makeatletter
\let\NAT@parse\undefined
\makeatother
\usepackage[colorlinks=true,linkcolor=red,citecolor=blue]{hyperref}

\let\oldtwocolumn\twocolumn
\renewcommand\twocolumn[1][]{%
    \oldtwocolumn[{#1}{
    \begin{center}
    \vskip-3ex
        \centering
        \includegraphics[width=\textwidth]{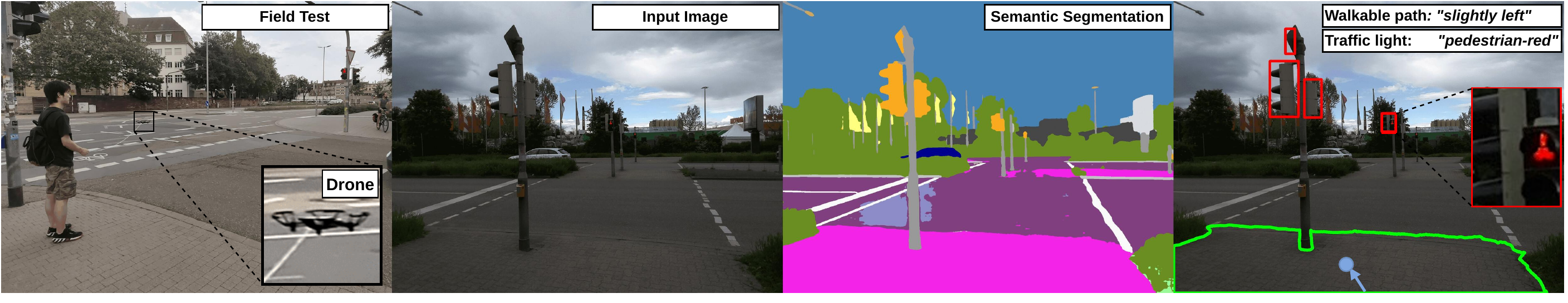}
        \vskip-2ex
        \captionof{figure}{\small Flying guide dog prototype. Images from left to right are: field test in front of an intersection, input image from the drone's perspective, semantic segmentation result, interpreted result with \textcolor{cyan}{centroid}, \textcolor{green}{boundary} and \textcolor{red}{bounding boxes} for the drone control algorithm.}
        \label{fig:banner}
    \end{center}
    }]
}

\begin{document}
\bstctlcite{IEEEexample:BSTcontrol}
    % Change to your title
    \title{\LARGE \bf
    Flying Guide Dog: Walkable Path Discovery for the Visually Impaired Utilizing Drones and Transformer-based Semantic Segmentation}
    
    \author{Haobin Tan$^{1}$, Chang Chen$^{1}$, Xinyu Luo$^{1}$,\\
    Jiaming Zhang$^{2}$, Constantin Seibold$^{2}$, Kailun Yang$^{2}$ and Rainer Stiefelhagen$^{2}$
    \thanks{This work was supported in part through the AccessibleMaps project by the Federal Ministry of Labor and Social Affairs (BMAS) under the Grant No. 01KM151112 and in part by the University of Excellence through the ``KIT Future Fields'' project. $^{1}$\{firstname.lastname\}@student.kit.edu, $^{2}$\{firstname.lastname\}@kit.edu (correspondence: Jiaming Zhang).}
    \thanks{Code and dataset will be made publicly available at: \url{https://github.com/EckoTan0804/flying-guide-dog}.}
    }
    \maketitle

    \begin{abstract}
        % Context / Problem
Lacking the ability to sense ambient environments effectively, \acrfull{bvip} face difficulty in walking outdoors, especially in urban areas. Therefore, tools for assisting \acrshort{bvip} are of great importance.
% Approach
In this paper, we propose a novel ``flying guide dog'' prototype for \acrshort{bvip} assistance using drone and street view semantic segmentation. Based on the walkable areas extracted from the segmentation prediction, the drone can adjust its movement automatically and thus lead the user to walk along the walkable path. By recognizing the color of pedestrian traffic lights, our prototype can help the user to cross a street safely. 
Furthermore, we introduce a new dataset named \acrfull{pvtl}, which is dedicated to traffic light recognition.
% Results & Impact
The result of our user study in real-world scenarios shows that our prototype is effective and easy to use, providing new insight into \acrshort{bvip} assistance.

    \end{abstract}

    \section{Introduction}
\label{sec:introduction}

    % More and more blind and visually impaired people
    It has been predicted that the number of \acrfull{bvip} will increase to more than 115 million by 2050 \cite{bourne2017magnitude}. 
    % Assistance is important
    Lacking the ability to sense ambient environments effectively, they face difficulty in walking outdoors, especially in urban areas. Therefore, assistive tools are indispensable.
    Previous electronic assistance systems focused on providing guiding information via audio \cite{article} or tactile feedback with wearable equipment \cite{7989772}. Although users can approach the destination based on these feedbacks, they need to repeatedly finetune their orientation during the navigation. 
    % Our approach
    To tackle this problem, we build a novel ``flying guide dog'' prototype, exploring the combination of drone and street view semantic segmentation. According to the user study conducted in \cite{avila2017dronenavigator}, drone navigation is more accurate and faster as it gives a continuous feedback in the direction of travel. Performing semantic segmentation on frames captured by the drone's camera, a variety of ambient visual information can be extracted, such as sidewalks, crosswalks, and traffic lights. Based on its perception of the environment, the drone adjusts itself %its movement 
    and leads the user to walk safely.
    To follow the drone, the user holds a string attached to the drone. Besides, the user receives voice prompts via a Bluetooth bone conduction headphone.
    %, as shown in the left image of Fig.~\ref{fig:banner}. 
    
    % Functions of our system
    % Walkable path discovery
    Discovering the walkable path is one of the major functions of our prototype.
    % One of the main functions of our prototype is the discovery of walkable paths. 
    Based on the segmentation prediction, walkable area (\textit{e.g.} sidewalks, crosswalks) can be discovered by their corresponding colors. In order to make the drone keep flying along the walkable path safely, we develop a control algorithm so that the drone can automatically adjust its direction and velocity according to the estimated centroid of the sidewalk.
    % Street crossing
    Another function is assisting the user to pass the pedestrian traffic light, \textit{i.e.} street crossing. Our prototype not only distinguishes pedestrian crossing lights from other types of traffic lights but also recognizes their color. Since there is currently no dedicated traffic light dataset containing both pedestrian and vehicle traffic lights, we introduce a new dataset called \textit{\acrfull{pvtl}}.  

    % User study
    To verify the effectiveness of our system, we further conduct a user study in real-world scenarios. The result indicates that our prototype is effective for visually impaired assistance and easy to use.
    
    In summary, our contributions are summarized as follows:
    \begin{compactitem}
        \item We propose a novel ``flying guide dog'' prototype utilizing the drone and semantic segmentation, which can effectively help \acrshort{bvip} to walk in urban scenarios safely.
        \item We develop a control algorithm to enable the drone to fly along the walkable path automatically and to interact with users via voice feedbacks in crossing streets.
        \item We introduce a new traffic light dataset named \textit{\acrfull{pvtl}}, which focuses on recognizing the category and the color of traffic lights simultaneously.
    \end{compactitem}
    
    \begin{figure*}[!t]
        \centering 
        \includegraphics[width=.9\textwidth]{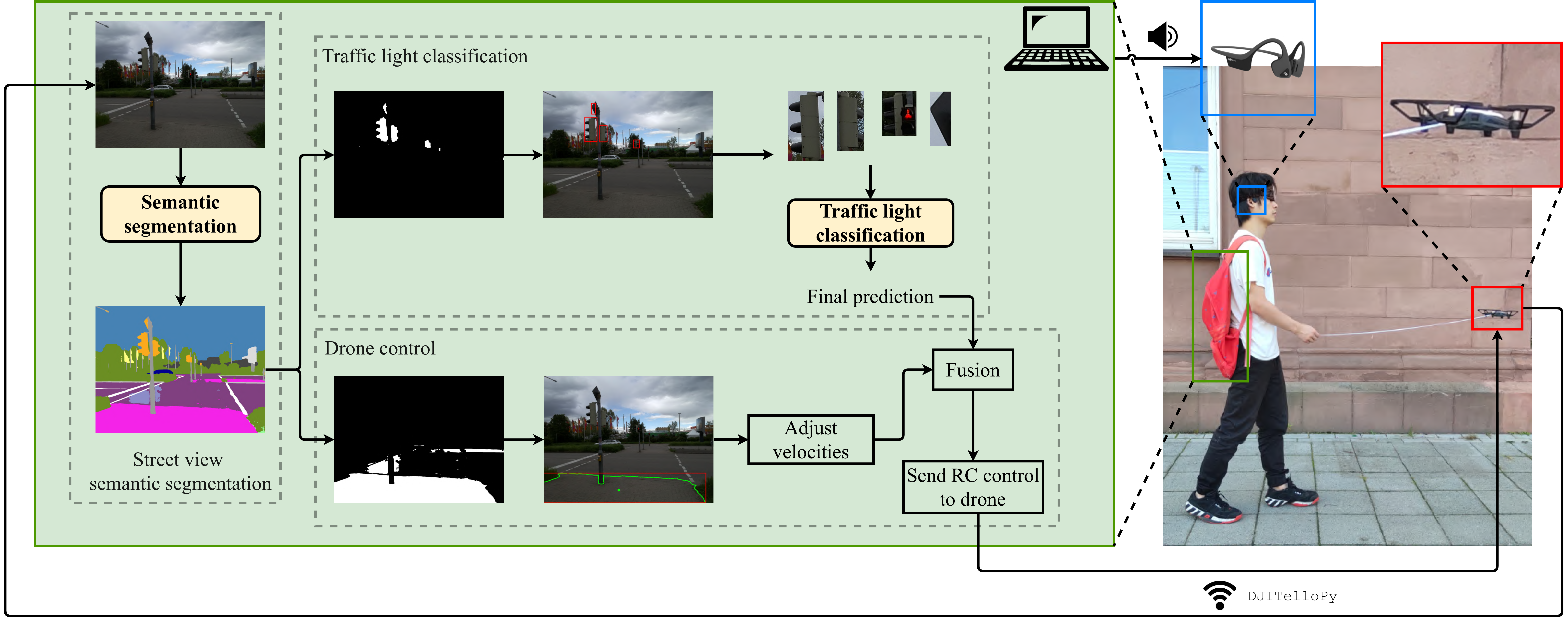}
        \vskip -1ex
        \caption{\small System overview. There are three components: street view semantic segmentation, traffic light classification, and drone control.
        }
        \vskip -3ex
        \label{fig:sys_structure}
    \end{figure*}

    \section{Related Work}
\label{sec:related_work}

    % \subsection{Assistive system}
    % \label{sec:related_work:assistive_system}
        \noindent \textbf{Assistive systems.} A variety of assistive devices dedicated for \acrshort{bvip} have been developed. 
        % Guide-dog robot system
        A guide-dog robot system, which makes use of template matching-based crossing light detection and Hough transform-based crosswalk detection, was designed for assisting the \acrshort{bvip} in self-walking \cite{Wei2014}.
        % real-time pedestrian crossing light detection algorithm
        In order to help \acrshort{bvip} cross the street, Cheng~\textit{et al.}~\cite{Cheng2018} proposed a real-time pedestrian crossing light detection algorithm for the \acrshort{bvip} using \acrfull{hog} and \acrfull{svm} and integrated it into a wearable navigation system. 
        % Intersection perception
        %Moreover,
        Yang~\textit{et al.}~\cite{Yang2018} unified intersection-centered perception tasks by utilizing real-time semantic segmentation. 
        While there were robotic guide dogs~\cite{chuang2018deep,xiao2021robotic} and drone-based assistive systems~\cite{avila2017dronenavigator,huppert2021guidecopter} for \acrshort{bvip}, we pioneer to realize path discovery and intersection navigation with semantic segmentation on drone videos.

    % \subsection{Semantic segmentation}
    % \label{sec:related_work:sem_seg}
        % FCN
        \noindent \textbf{Semantic segmentation.} As a seminal work, FCN \cite{long2015fully} performs dense pixel-to-pixel classification in an end-to-end manner.
        %Subsequent
        Subsequent networks follow the per-pixel recognition paradigm and significantly improve scene segmentation performance.
        % PSPNet
        Introducing the global pyramid pooling feature, PSPNet \cite{zhao2017pyramid} embeds multi-scale scenery context features in an FCN-based pixel prediction framework.
        % Attention- and MLP-based methods
        Recently, non-local attention layers and \acrfullpl{mlp} are frequently used for dense prediction tasks like image semantic segmentation~\cite{chen2021cyclemlp,zhang2021trans4trans}.
        % SegFormer
        In contrast to FCN-based models, SegFormer \cite{xie2021segformer} combines a hierarchical Transformer \cite{Vaswani2017} encoder and a lightweight All-\acrshort{mlp} decoder. It yields a powerful representation without complex and computationally demanding modules.
        % Our
        Compared to existing systems that use FCN-based semantic segmentation for assisting \acrshort{bvip}, we build our flying guide dog system with SegFormer, which can yield an efficient and robust segmentation.

     \section{Approach}
 \label{sec:approach:approach}
    
    % Goal
    Our goal is to implement a ``flying guide dog'' prototype, which is able to automatically discover walkable areas and avoid obstacles so that it can guide the \acrshort{bvip} person to walk safely.
    % Main procedure
    The drone is connected to the computer using \texttt{DJITelloPy} \footnote{\url{https://github.com/damiafuentes/DJITelloPy}} library via Wifi. 
    % Semantic segmentation
    For each frame captured by the drone's camera, the semantic segmentor outputs a colorized prediction. 
    % Drone control
    To make the drone fly along the walkable path, the largest walkable area is extracted and its centroid is then estimated. On the basis of centroid estimation, velocity adjustment is computed. 
    % Traffic light classification
    Meanwhile, if traffic lights are detected, we crop them out and input them into the traffic light classification model.  
    % Fusion
    Fusing the classification prediction and velocity adjustment, a \acrfull{rc} command is sent to the drone. Additionally, the user gets voice prompts via a Bluetooth-connected bone conduction headphone.
    
    As illustrated in Fig. \ref{fig:sys_structure}, our system is composed of three main modules: 
    \begin{enumerate*}[label=(\alph*)]
        \item {street view semantic segmentation} (Sec.~\ref{sec:approach:sem_seg}),
        \item {traffic light classification} (Sec.~\ref{sec:approach:traffic_light_cls}),
        and \item {drone control} (Sec.~\ref{sec:approach:drone_control})
    \end{enumerate*}.

    \subsection{Semantic segmentation}
    \label{sec:approach:sem_seg}
    
        \noindent \textbf{Robustness and effeciency.}
        A vital requirement for the segmentation model is real-time performance. According to the comparison of real-time semantic segmentation models,
        SegFormer-B0 \cite{xie2021segformer} has significant advantages on speed and accuracy. Futhermore, it is more robust to common corruptions and perturbations.
        
        \noindent \textbf{Datasets}.
        While Cityscapes~\cite{cordts2016cityscapes} are recorded in a unified setting,
        Mapillary Vistas~\cite{neuhold2017mapillary} are globally taken by diverse devices from different viewpoints. Moreover, comprising 25,000 densely annotated street level images into 66 categories, Mapillary Vistas is 5$\times$ larger than Cityscapes in terms of fine-grained annotations. Therefore, Mapillary Vistas is more diverse and promising for yielding robust models, thus suitable for our prototype.
        
        \noindent \textbf{Implementation details}.
        We train SegFormer-B0 \cite{xie2021segformer} on Mapillary Vistas \cite{neuhold2017mapillary} using \texttt{mmsegmentation} \footnote{\url{https://github.com/open-mmlab/mmsegmentation}} codebase on a single GTX 1080 Ti GPU. During training, we apply data augmentation techniques, including random resizing with ratio 0.5-2.0, random horizontal flipping, and random cropping to $768 \times 768$. We train the model for 160K iterations with AdamW optimizer. Initial \acrlong{lr} is set to $6E-5$  
        and a polynomial \acrlong{lr} decay scheduler with factor 1.0 is employed, following the training strategy in \cite{xie2021segformer}.
        
    \begin{figure}[!t]
      \centering 
      \includegraphics[width=0.7\columnwidth]{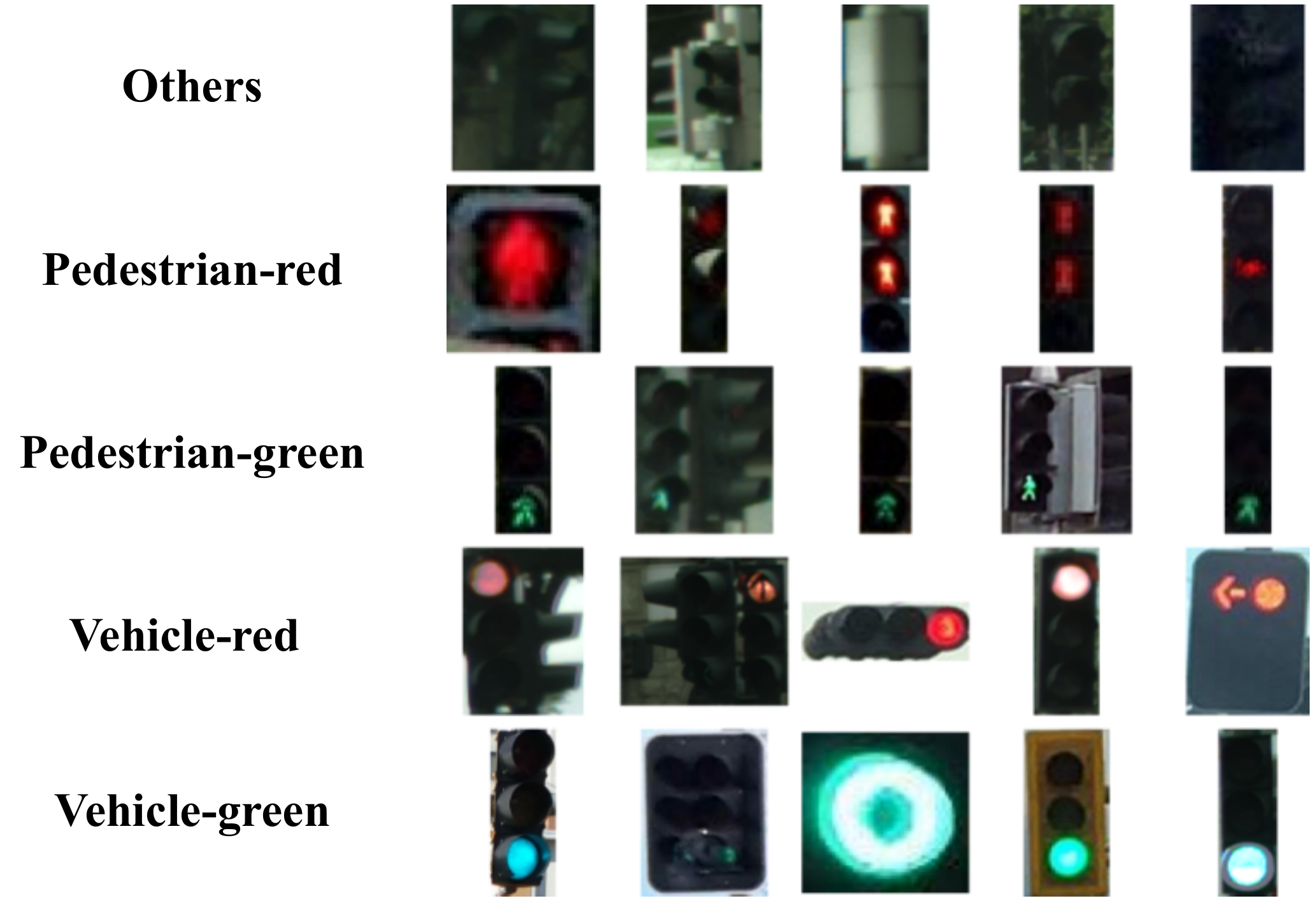}
    %   \vskip -1ex
      \caption{\small \acrshort{pvtl} dataset with 5 categories. The ``others" refers to side or back of traffic lights and unilluminated traffic lights.}
      \label{fig:traffic_light_eg_imgs}
      \vskip -4ex
    \end{figure}
    
    \subsection{Traffic light classification}
    \label{sec:approach:traffic_light_cls}
        
        After semantic segmentation, a prediction mask regarding traffic lights can be obtained. Based on the mask, the traffic light patch can be cropped from its full-resolution image. Performing image classification in a cropped patch is more computationally efficient than in a full-scale image. To simplify and unify the recognition of the color and category of traffic lights, we train a light CNN, which is expected to be more accurate and efficient. More importantly, it should eliminate the influence of vehicle traffic lights. Yet, currently there is not a dataset directly suitable for our task.
        
        \noindent \textbf{\acrshort{pvtl} dataset}.
            To facilitate traffic light classification, we introduce a new dataset called \textit{\acrfull{pvtl}} with the goal of distinguishing pedestrian traffic lights from vehicle traffic lights. We perform four steps to collect this dataset. Firstly, we crop the traffic lights from Cityscapes \cite{cordts2016cityscapes}, Mapillary Vistas \cite{neuhold2017mapillary}, and Pedestrian lights \cite{roters2011recognition} based on their annotations. Secondly, we clean up those image patches with resolution smaller than $8\times 8$. Then, we manually annotate the images into $5$ categories: \{\emph{pedestrian-red, pedestrian-green, vehicle-red, vehicle green, others}\}. Last but not least, a class balancing is performed to maintain 300 images for each class. We visualize some representative examples in Fig. \ref{fig:traffic_light_eg_imgs}. The category \emph{others} refers to the side and back of the traffic lights or traffic lights that are not illuminated. All data and annotations will be made publicly available.
        
\begin{figure}[!t]
          \centering 
          \includegraphics[width=\columnwidth]{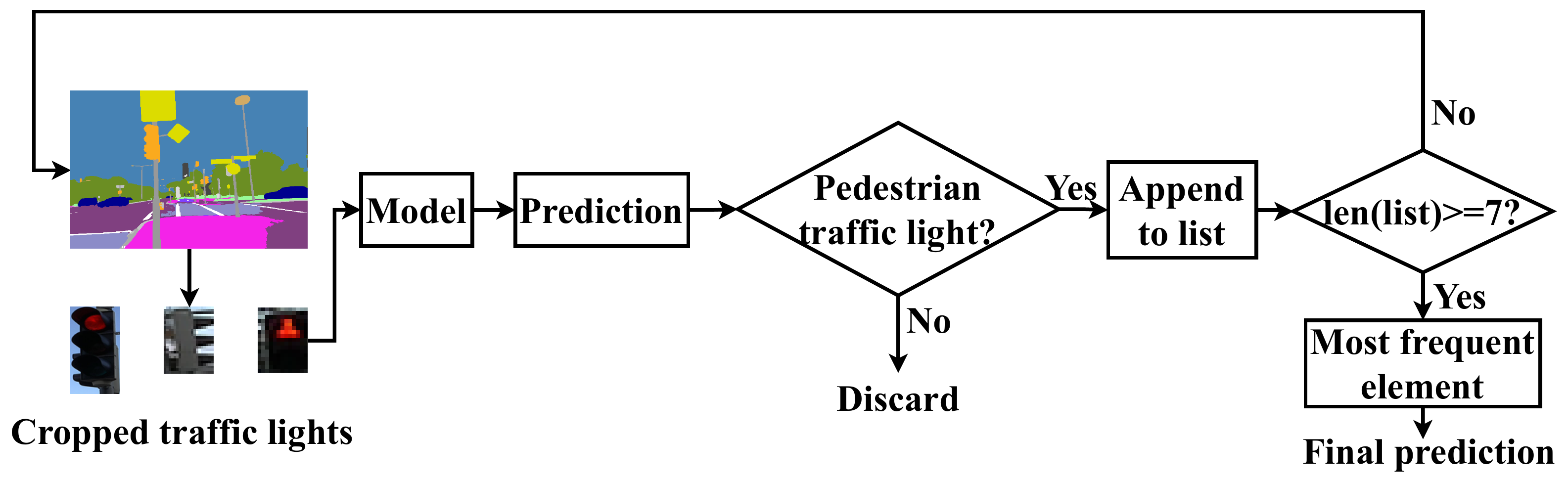}
          \caption{Multi-frame based traffic light classification.}
          \label{fig:algorithm}
        \end{figure}

        \noindent \textbf{Pedestrian light prediction}.
        As shown in Fig. \ref{fig:algorithm}, the input of the model is the traffic light patch cropped from every single frame when crossing traffic light intersections. In order to eliminate the influence of vehicle traffic lights, we keep only the pedestrian light prediction and discard others. Each prediction will be appended to the buffer list, whose length is 7 frames. Based on the accumulated predictions, the most frequent class is output as the final traffic light prediction, which can robustify the system and ensure the safety. 
    
    \subsection{Drone control}
    \label{sec:approach:drone_control}
\begin{figure}[!b]
              \centering 
              \includegraphics[width=\columnwidth]{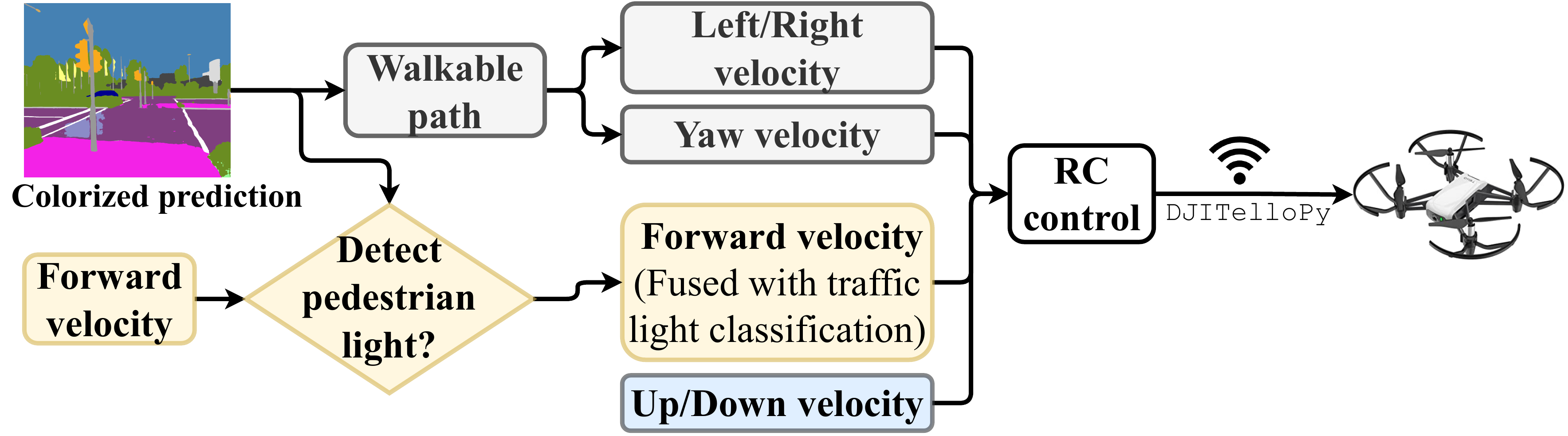}
              \caption{\small Drone control overview. Four types of  velocity are updated based on the walkable path and the traffic light prediction.}
              \label{fig:control_overview}
        \end{figure}

        After achieving semantic segmentation and traffic light classification, four types of velocity are calculated to control the drone, as depicted in Fig. \ref{fig:control_overview}. Among them, the up/down velocity $v_{ud}$ is obtained by the Tello’s vision positioning system to maintain the flying height $h_{target}$, which is preset as $1.2m$ to ease the user interaction.
        Finally, \acrshort{rc} commands are sent to the drone using \texttt{DJITelloPy}. 
        A detailed description of the control strategy is presented in Algorithm \ref{algo:drone_control}. 
        
        \begin{algorithm}[!ht]
            \small
            \SetAlgoLined
            \SetKwInOut{Input}{input}
            \SetKwInOut{Output}{output}
            \Input{colorized prediction $C \in \mathbb{R}^{H\times W \times 3}$,
                pedestrian crossing light $color \in \{red, green, None\}$, \\
                current altitude $h$}
            \Output{up/down velocity $v_{ud}$, yaw velocity $v_{yaw}$, left/right velocity $v_{lr}$, forward velocity $v_{f}$}
            \BlankLine
            Initialize parameters: pixel threshold $\theta_{conf}$, target altitude $h_{target}$, speed up $speedup$, up/down velocity $v_{ud}$, forward velocity $v_{f,0}$;\ lists: preset yaw velocities $list_{yaws}$, \ binary control code $list_{codes}$\;
            $start\_crossing \leftarrow \text{false}$\;
            \While{drone is flying and video stream is on}{
                \tcp{Maintain target altitude}
                \If{$h \neq h_{target}$}
                    {$v_{ud}$ $\leftarrow$ $(h_{target} - h) / h_{target} * v_{ud}$\;}
                \BlankLine
                \tcp{Fly along walkable path}
                Extract largest walkable area $L_{walkable}$ from $C$\;
                $(x_{centroid}, y_{centroid}) \leftarrow \operatorname{estimate\_centroid}(L_{walkable})$ \;
                $v_{lr} \leftarrow x_{centroid} - \frac{W}{2}$\;
                $(R_{l}, R_{m}, R_{r})\leftarrow \operatorname{partition}(L_{walkable})$\;
                \For{$p \in (R_{l}, R_{m}, R_{r})$}{
                    $conf \leftarrow \operatorname{mean}(p)$\;
                    $code \leftarrow \operatorname{binary\_conversion}(conf, \theta_{conf})$\;
                    Append $code$ to $list_{codes}$\;
                }
                $v_{yaw} \leftarrow \operatorname{get\_yaw\_vel}(list_{yaws}, list_{codes})$\;
                \BlankLine
                \tcp{Fusion with traffic light classification}
                \uIf{$color = None$}{
                    $v_{f}$ $\leftarrow$ $v_{f,0}$\;
                }
                \uElseIf{$color =green$ or $start\_crossing$}{
                    $v_{f} \leftarrow v_{f,0} + speedup$\;
                    $start\_crossing \leftarrow \text{true}$\;}
                \Else{
                    $v_{f}\leftarrow0$\;
                }
            }
            %\vskip -4ex
            \caption{Drone control}
            \label{algo:drone_control}
            
        \end{algorithm}
        \setlength{\textfloatsep}{6pt}
    
        \noindent \textbf{Fly along walkable path}.
        Based on the mask of the walkable path (\textit{e.g.} \emph{sidewalks} and \emph{crosswalks}), the largest walkable area $L_{walkable}$ is cropped from the colorized prediction $C$. Through the image moment of $L_{walkable}$, the centroid of the walkable area is computed. Smoothing the position of the contour centroid with high-pass and low-pass filters makes the movement more stable. Horizontal difference between the filtered centroid and the center of the whole image is used as the reference of the left/right velocity $v_{lr}$. The horizontal adjustment is important to prevent most of collisions and assist \acrshort{bvip} to walk safely. After adjusting the left/right velocity, the drone can constantly fly in the middle of the walkable path.
        
        In order to fly towards the direction of a walkable path, the yaw velocity $v_{yaw}$ is required to veer. Motivated by \cite{9302760}, the image with the largest walkable area is further divided into three partitions horizontally. Then the confidence $conf$ for each part, the mean of the pixel values, is compared with a pre-adjusted threshold $\theta_{conf}$ to obtain the final binary control code $list_{codes}$. Each control code is assigned to separate yaw speed $v_{yaw}$ in $list_{yaws}$.
    
        \noindent \textbf{Fusion with traffic light classification}.
        When encountering a traffic light during walking, the drone control with a preset forward velocity $v_{f,0}$ will be updated and integrated with the traffic light recognition.
        If the pedestrian traffic light turns green, the drone accelerates forward. If it is red, the forward velocity is set to 0 and the drone will hover in place until the light turns green. If the light changes from green to red during crossing, the drone will maintain the adjusted velocity.
        
        \noindent \textbf{Voice feedback.}
        At each step of traversing the pedestrian traffic lights, voice prompts (\textit{i.e.} speech) will be sent to the user through the bone conduction headset, such as ``stop'' at the red light and ``go'' at the green light. Using voice feedback can always keep a certain safe distance between the user and the drone.

    \section{Experiments}
\label{sec:experiments}

    \subsection{Semantic segmentation results}
    
\begin{figure}[ht]
            \centering 
            \begin{subfigure}[b]{0.5\columnwidth}
                \centering
                \includegraphics[width=.9\columnwidth]{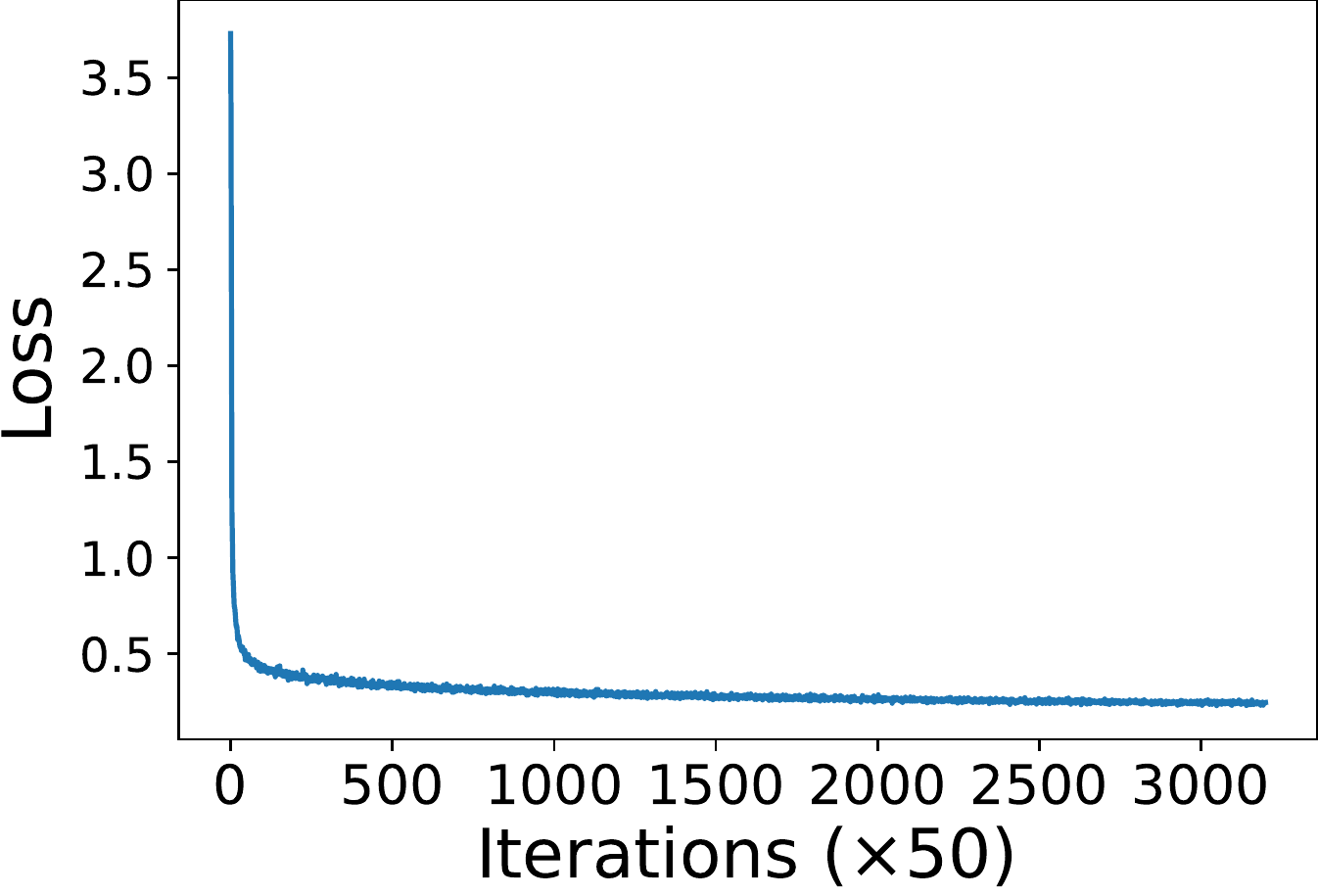}
                \caption{Loss}
                \label{fig:train_loss}
            \end{subfigure}%
            \hfill
            \begin{subfigure}[b]{0.5\columnwidth}
                \centering
                \includegraphics[width=.9\columnwidth]{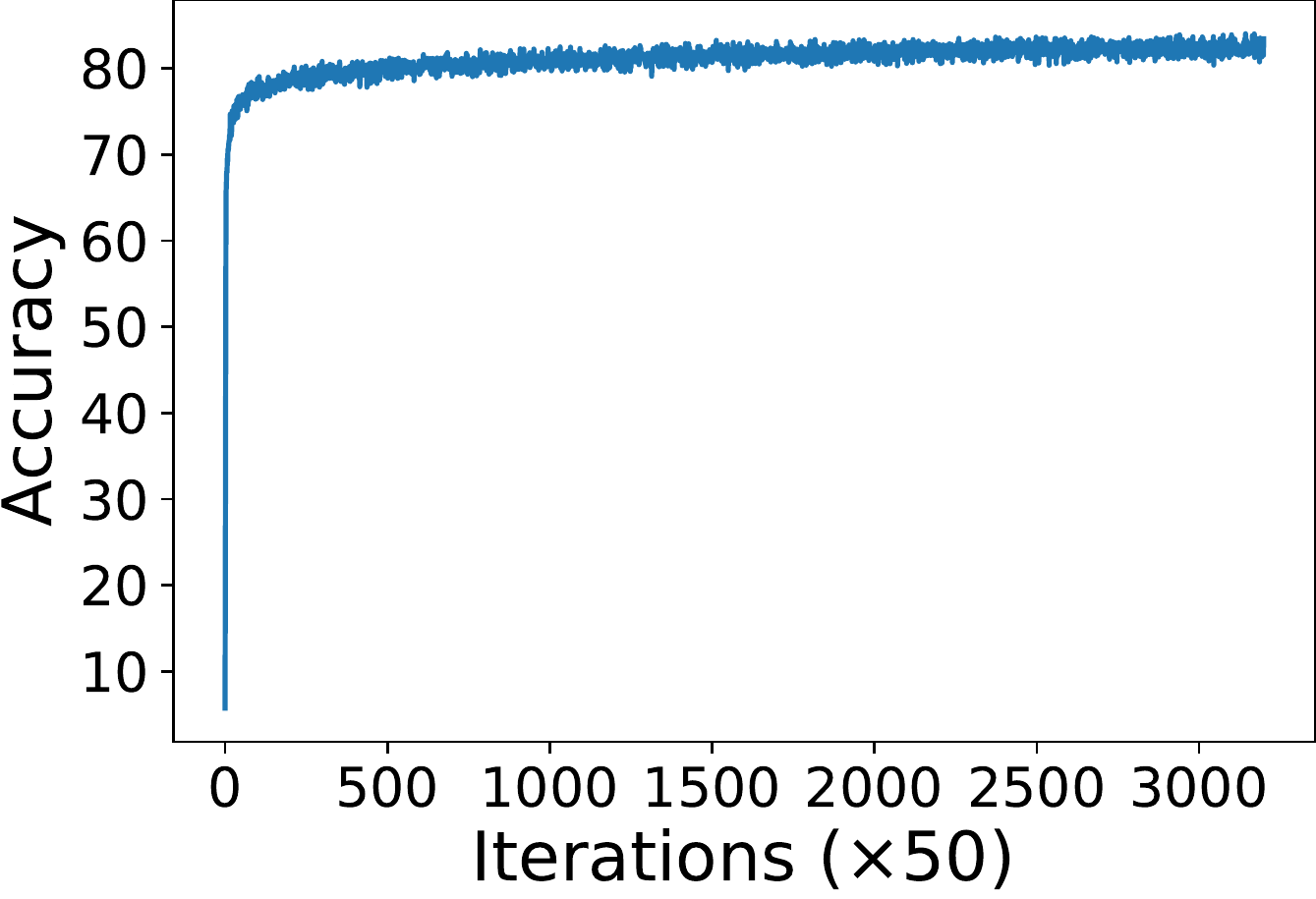}
                \caption{Accuracy}
                \label{fig:train_acc}
            \end{subfigure}
            \vskip -2ex
            \caption[]{\small Training Loss (left) and accuracy (right) of SegFormer-B0 \cite{xie2021segformer} on Mapillary Vistas \cite{neuhold2017mapillary}.}
            \label{fig:train_log}
        \end{figure}

        \noindent \textbf{Overall result}. To investigate the training procedure of SegFormer-B0 \cite{xie2021segformer} on Mapillary Vistas \cite{neuhold2017mapillary}, we present the logging of training loss in Fig.~\ref{fig:train_loss} and the one of accuracy in Fig.~\ref{fig:train_acc}, which are measured in every 50 iterations. In the first few hundred iterations, the loss drops dramatically. After 25,000 iterations, the loss and accuracy start to converge and remain relatively steady. After training for 160K iterations, SegFormer-B0 yields 41.83\% \acrfull{miou} on validation set of Mapillary Vistas, with the shorter side of input image being 1024.
        
        \noindent \textbf{Per-class result}. \acrfullpl{iou} of some categories of interest for \acrshort{bvip} and related to our task are reported in Table \ref{tab:mapillary_cls_iou}.  
        SegFormer-B0 \cite{xie2021segformer} achieves compelling results on categories that are significant for walkable area discovery, such as \textit{Road}~(85.20\%), \textit{Sidewalk}~(62.32\%), and \textit{Land Marking - Crosswalk}~(63.80\%), which guarantee the accuracy and effectivness of our system. Moreover, the model obtains 61.06\% accuracy of the \emph{Traffic Light}, thus it can generate a sufficient patch for the downstream traffic light classification. 
    
        \begin{table}[!t]
            \centering
            \caption{\small Per-class \acrshort{iou} of SegFormer-B0 \cite{xie2021segformer} on Mapillary Vistas \cite{neuhold2017mapillary} validation set in $1024\times 2048$ resolution.}
            \resizebox{.5\columnwidth}{!}{%
                \begin{tabular}{|l|c|}
                    \hline
                    \textbf{Category}        & \textbf{IoU}  \\ \hline
                    Bike Lane                & 31.87         \\ \hline
                    Crosswalk - Plain        & 21.09         \\ \hline
                    Curb                     & 53.20         \\ \hline
                    Pedestrian Area          & 31.24         \\ \hline
                    Road                     & 85.20         \\ \hline
                    Sidewalk                 & 62.32         \\ \hline
                    Lane Marking - Crosswalk & 63.80         \\ \hline
                    Traffic Light            & 61.06         \\ \hline
                    Traffic Sign (Front)     & 68.67         \\ \hline
                    Car                      & 86.84         \\ \hline
                    % \textbf{Average~(mIoU)}  & -             \\ \hline
                \end{tabular}%
            }
            \label{tab:mapillary_cls_iou}
        \end{table}
        
        \noindent \textbf{Comparison to state-of-the-art methods.}
        Based on the evaluation on Mapillary Vistas \cite{neuhold2017mapillary}, Table \ref{tab:mapillary_sota} summarizes results including \acrfull{fps}, \acrshort{miou}, and parameters. Compared to other real-time approaches (\textit{e.g.} RGPNet \cite{arani2021rgpnet}), SegFormer-B0 \cite{xie2021segformer} achieves comparable accuracy with significantly fewer parameters, which shows great potential for deployment on edge devices and mobile devices. Note that to propose a novel network is not our scope in this work, instead to introduce the feasibility of deploying a Transformer-based semantic segmentation model on the assistive system. Besides, our system can be constructed by arbitrary efficient backbones.
        
        \begin{table}[ht]
        \centering
        \caption{\small Comparison of \acrlong{sota} methods on Mapillary Vistas \cite{neuhold2017mapillary} validation set. Inference speed (\acrshort{fps}) is calculated on $1024\times 2048$ image resolution. }
        \label{tab:mapillary_sota}
        \resizebox{.9\columnwidth}{!}{%
            \begin{tabular}{|c|l|ccc|}
                \hline
                & \textbf{Method}     & \textbf{\acrshort{fps}} $\uparrow$          & \textbf{\acrshort{miou}} (\%) $\uparrow$      & \textbf{Params (M)} $\downarrow$ \\ \hline
                {\multirow{6}{*}{\rotatebox[origin=c]{90}{\scriptsize \textit{Non Real-Time}}}} 
                &BiSeNet (R101)~\cite{yu2018bisenet}       & 9.3           & 20.4          & 50.1 \\ 
                &TASCNet (R50)~\cite{li2018tascnet}        & \textbf{11.9}          & 46.4          & \textbf{32.8} \\
                &TASCNet (R101)~\cite{li2018tascnet}       & 8.8           & 48.8          & 51.8 \\
                &ShelfNet (R101)~\cite{zhuang2019shelfnet}      & 9.1           & 49.2          & 57.7 \\
                &RGPNet (R101)~\cite{arani2021rgpnet}        & 10.8          & 50.2          & 52.2 \\
                &RGPNetB(WRN38)~\cite{arani2021rgpnet}      & 3.4           & \textbf{53.1}          & 215.0  \\
                \hline
                {\multirow{3}{*}{\rotatebox[origin=c]{90}{\tiny \textit{Real-Time}}}} 
                &RGPNet (HarDNet39D)~\cite{arani2021rgpnet}  & 34.7          & \textbf{42.5}          & 9.4   \\
                &RGPNet (R18)~\cite{arani2021rgpnet}         & \textbf{35.7}          & 41.7          & 17.8  \\
                &SegFormer-B0 (ours) & 15.2             & 41.8          & \textbf{3.8}   \\
                \hline
            \end{tabular}%
        }
        \end{table}
    
    \subsection{Qualitative comparison}
    \label{sec:experiments:qualitative_comparison}
\begin{figure}[!t]
          \centering 
          \includegraphics[width=\columnwidth]{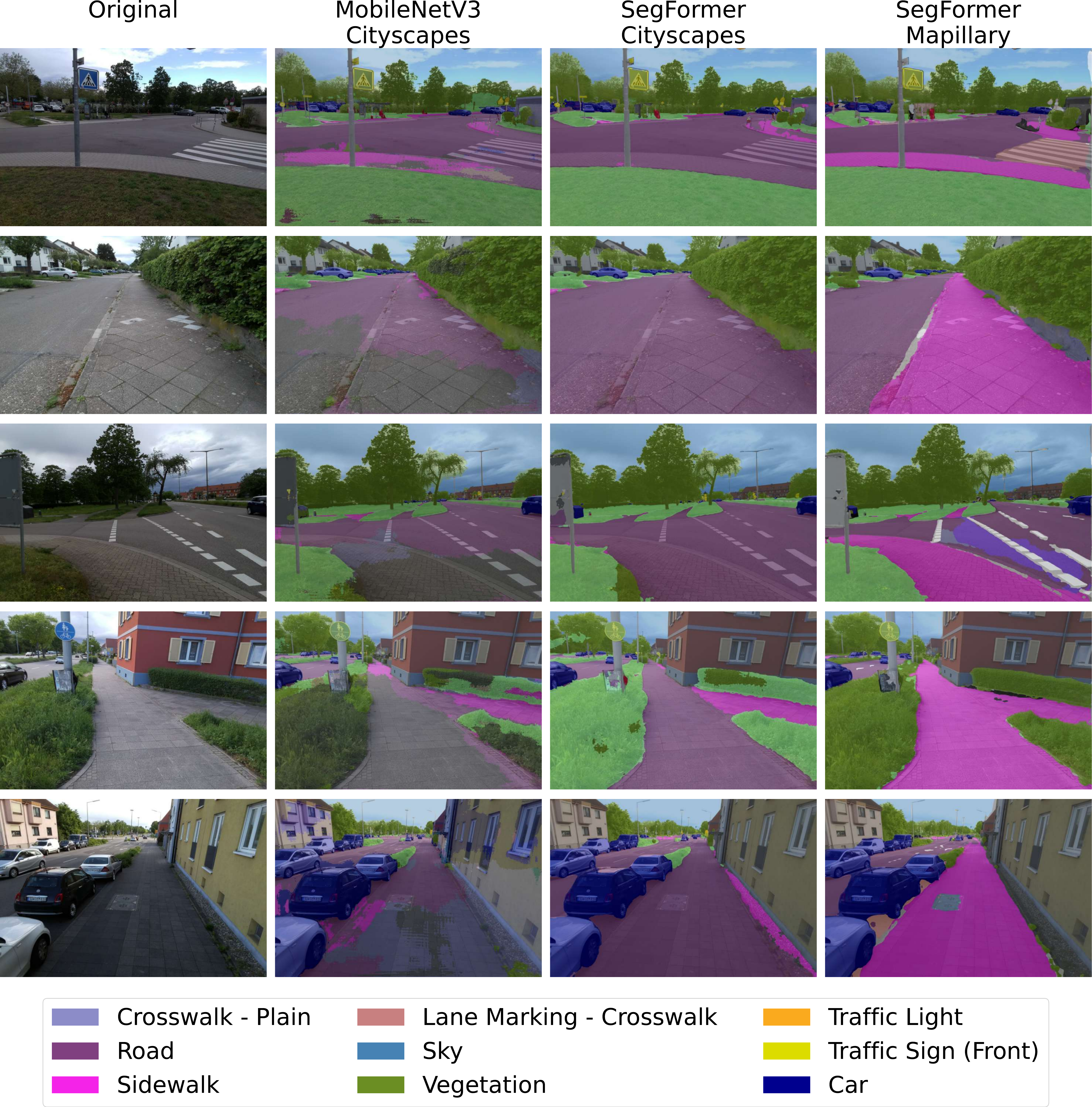}
          \caption{\small Qualitative results of predictions of different models trained on different datasets. 
          Compared to models trained on Cityscapes \cite{cordts2016cityscapes}, SegFormer-B0 \cite{xie2021segformer} trained on Mapillary Vistas \cite{neuhold2017mapillary} segments sidewalk and crosswalk much better.}
          \label{fig:seg_model_comparison}
          \vskip -1ex
        \end{figure}

        Fig. \ref{fig:seg_model_comparison} depicts the qualitative results.
        For images taken from the pedestrian's \acrfull{pov}, models trained on Cityscapes is almost unable to segment the sidewalk from the road. 
        This result may be explained by the fact that images of Cityscapes \cite{cordts2016cityscapes} are taken by a camera mounted behind the vehicle's windshield, \textit{i.e.}, from the vehicle's \acrshort{pov}. Accordingly, the models lack the ability of generalization to segment images from the viewpoints of the pedestrian. 
        In contrast, SegFormer-B0 trained on Mapillary Vistas \cite{neuhold2017mapillary} outputs a clearly better segmentation between sidewalks, crosswalks, and roads, thanks to diverse viewing perspectives in the training samples. 
    
    \subsection{Traffic light classification results}
    \label{sec:experiments:tl_cls}
        As shown in Table \ref{tab:model_comparision}, we first construct our light CNN model which contains 5 convolutional layers and 3 fully connected layers. In addition, we fine-tune ResNet-18 \cite{he2016deep} as our alternative. Both models are trained on \acrshort{pvtl} dataset for 25 epochs. Compared with ResNet-18, our light CNN model is slightly less accurate but more lightweight and computationally efficient. 
        \begin{table}[htb]
            \centering
            %\small
            \caption{\small Comparison of traffic light classification models.}
            \resizebox{0.9\columnwidth}{!}{%
            \begin{tabular}{|l|c|c|}
                \hline
                \textbf{Models} & ResNet-18 \cite{he2016deep} & Light CNN \\ \hline
                \textbf{Weight Initialization} & Pretrained on ImageNet & Default \\ \hline
                \textbf{Accuracy} $\uparrow$ & \textbf{90\%} & 83\% \\ \hline
                \textbf{FPS} $\uparrow$ & 167 & \textbf{670} \\ \hline
                \textbf{\#Params (M)} $\downarrow$ & 11.6 & \textbf{2.4} \\ \hline
                \textbf{Flops (GFLOPs)} $\downarrow$ & 7.2 & \textbf{0.35} \\ \hline
            \end{tabular}%
            
            }
            \label{tab:model_comparision}
        \end{table}
    
    \subsection{Runtime analysis}
    \label{sec:experiments:runtime}
        % Hardward
        The drone we adopt for runtime measurement is DJI Tello. Equipped with a RGB camera, it supports 720p HD transmission. To handle image processing, model inference, and further drone control, we utilize a laptop with an Intel i7 CPU and an Nvidia GeForce GTX 1050 Ti GPU. After warming up for 200 frames, we measure and average the runtime of the subsequent 800 frames. As summarised in Table~\ref{tab:runtime_performance}, the mean processing time of a single frame at the resolution of $320 \times 240$ is $85 ms$, mostly on semantic segmentation ($71 ms$). In other words, the speed of our prototype is 11.7 FPS.%\acrfull{fps}.

        \begin{table}[ht]
            \centering
            \caption{\small Running time analysis.
            ``Others'' refers to the time that costs for frame transmission, velocity adjustment, and sending \acrshort{rc} command.}
            \resizebox{.8\columnwidth}{!}{%
            \begin{tabular}{|l|c|c|}
                \hline
                \textbf{Models} & \begin{tabular}[c]{@{}c@{}}SegFormer \cite{xie2021segformer} + \\ ResNet-18 \cite{he2016deep}\end{tabular} &
  \begin{tabular}[c]{@{}c@{}}SegFormer \cite{xie2021segformer} + \\ Light CNN\end{tabular} \\ 
                \hline
                \textbf{Segmentation (ms)} $\downarrow$                & \textbf{70}                             & 71                             \\ \hline
                \textbf{Classification (ms)} $\downarrow$  & 18                             & \textbf{14}                             \\ \hline
                \textbf{Others (ms)} $\downarrow$                       & 1                              & 1                              \\ \hline
                \textbf{All (ms)} $\downarrow$                           & 89                             & \textbf{85}                             \\ \hline
                \textbf{FPS} $\uparrow$                            & 11.274                         & \textbf{11.713}                         \\ \hline
            \end{tabular}%
            }
            \label{tab:runtime_performance}
        \end{table}
    
    \subsection{User study}
    \label{sec:experiments:user_study}
    To evaluate the assistance functions of our system, an user study in real world scenarios is conducted using NASA Task Load Index (NASA-TLX) method \cite{hart2006nasa}. 
    
\begin{figure}[b]
            \centering 
            \begin{subfigure}[b]{0.3\columnwidth}
                \centering
                \includegraphics[width=0.9\columnwidth]{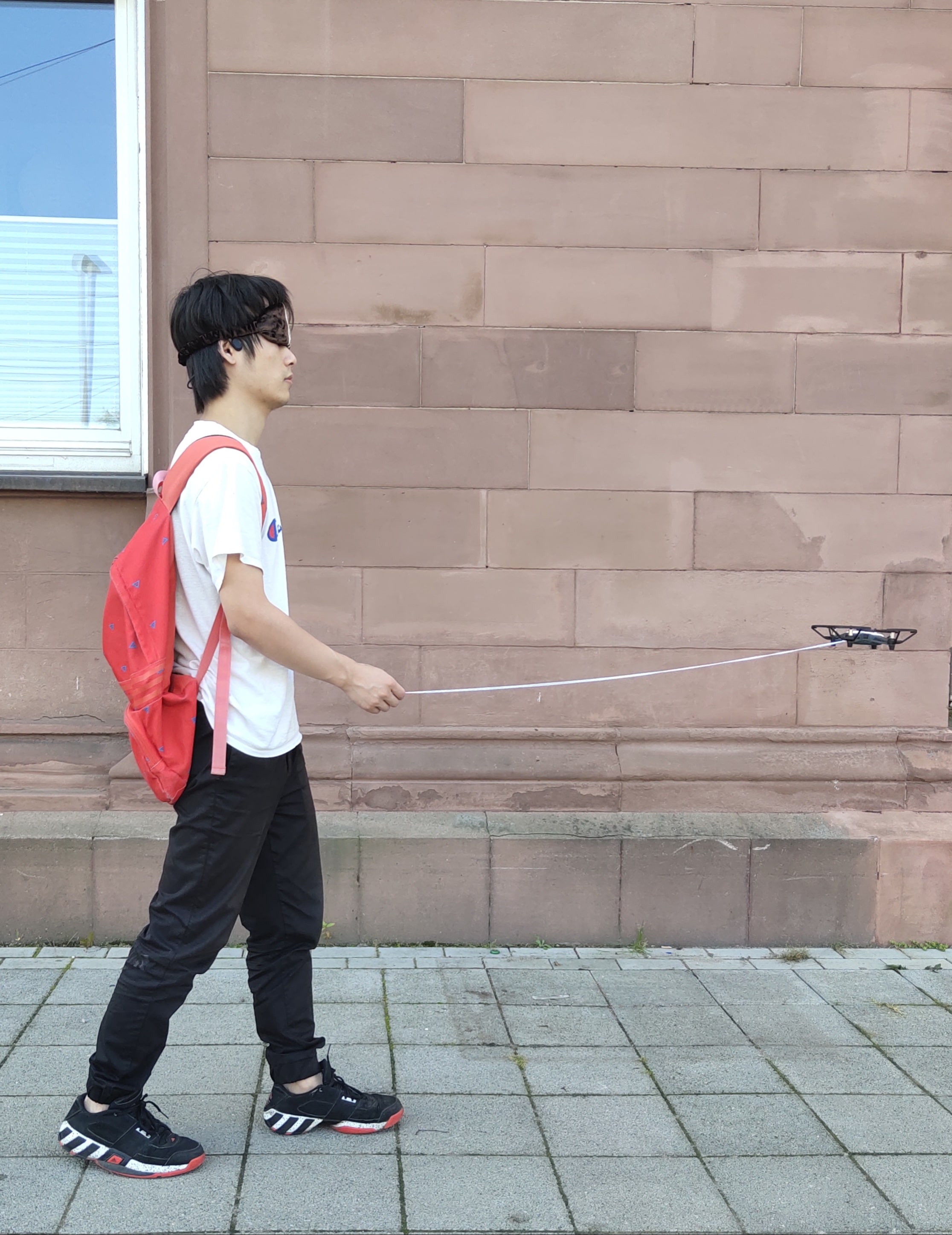}
                \caption{Sidewalk}
                \label{fig:user_study_sidewalk}
            \end{subfigure}%
            \hfill
            \begin{subfigure}[b]{0.7\columnwidth}
                \centering
                \includegraphics[width=0.9\columnwidth]{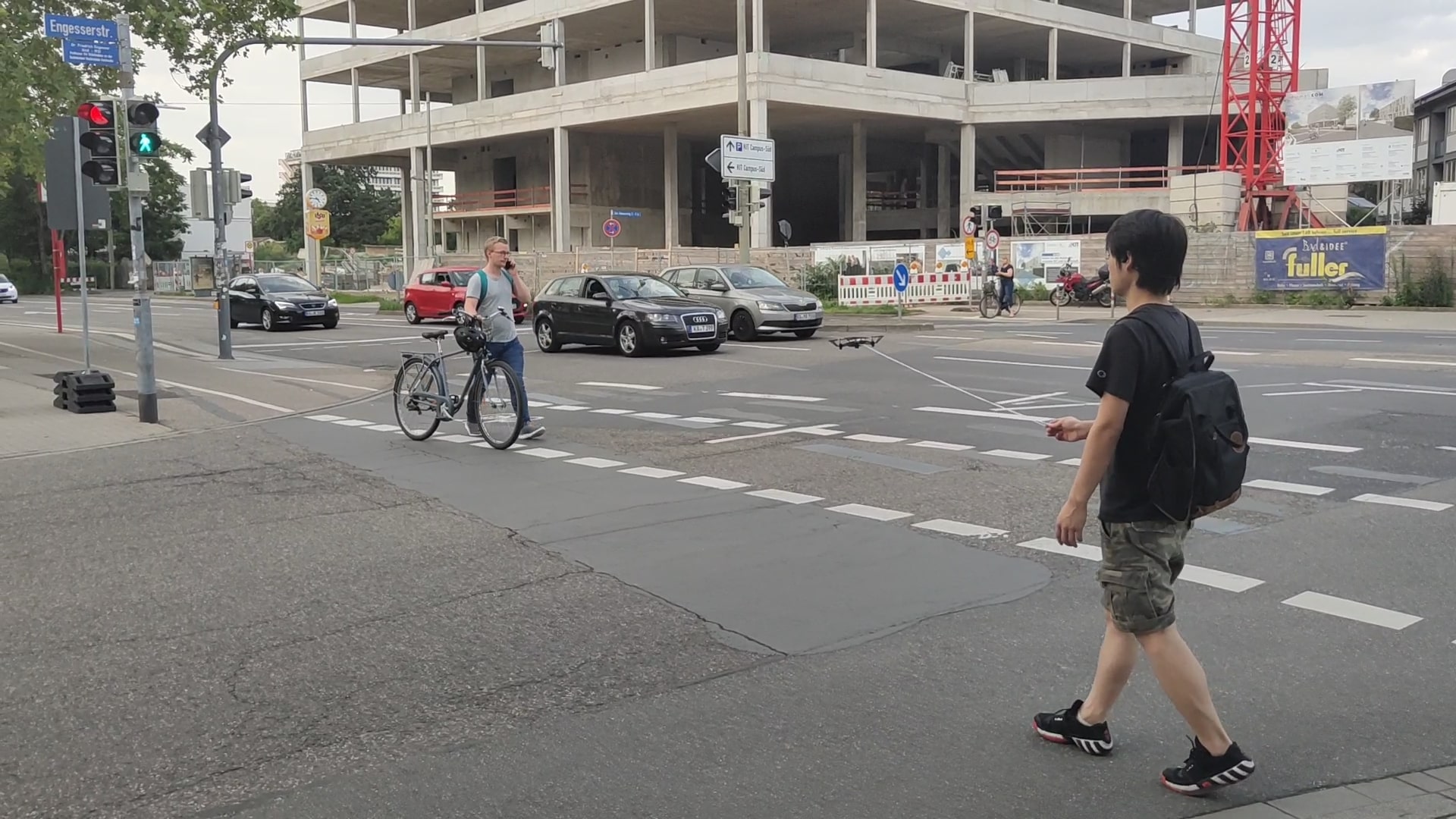}
                \caption{Street crossing}
                \label{fig:user_study_crosswalk}
            \end{subfigure}
            \vskip -1ex
            \caption[]{\small User study consists of two scenarios:
                \begin{enumerate*}[label=(\alph*)]
                    \item sidewalk
                    \item street crossing
                \end{enumerate*}.
                To simulate visually inpaired people, the participant wears a blindfold during the user study.
            }
            \label{fig:user_study}
        \end{figure}

    \noindent \textbf{User study setup}. During the user study, we use the same hardware setup as in Section \ref{sec:experiments:runtime}. The laptop is placed in a backpack, as shown in Fig. \ref{fig:user_study_sidewalk}.
    Six participants aged between 24 and 35, including 5 males and 1 female, are sighted but blindfolded during the test to simulate \acrshort{bvip}. The drone flies in front of the participants to simulate the real guide dog and to get an unobstructed view of the environment. The participant follows the drone by feeling the traction from a string attached to the drone. 
    
    Our prototype is evaluated in two open-space scenarios, as exhibited in Fig. \ref{fig:user_study}. The first scenario focuses on walking along the sidewalk in $20$ meters length. Some obstacles, such as \emph{bicycles} and \emph{pedestrians}, were randomly appeared in the sidewalk to test the obstacle avoiding function. The second scenario, an intersection including vehcile and pedestrian traffic lights, is selected to test the street crossing functionality according to the pedestrian light prediction. After testing, participants fill multi-aspected questionnaires based on the six categories listed in Table \ref{tab:questionnaires}. They score each aspect from 1 to 10, among which 1 and 10 mean \emph{strongly disagree} and \emph{strongly agree}, respectively.
    \begin{table}[t]
        \centering
        \small
        \caption{\small Six aspects of the user study questionnaire.}
        \begin{tabular}{|p{0.2\columnwidth}|p{0.7\columnwidth}|}
            \hline
            \textbf{Aspects}& \textbf{Statements}\\
            \hline
            Orientation& I can walk in the proper direction.\\
            \hline
            Position& I can walk in the middle of the path.\\ 
            \hline
            Traffic Light& I can cross the road at the right time according to the pedestrian lights.\\ 
            \hline
            Learnability& I can get familiar with this system easily.\\
            \hline
            Mentally easy to use& I don't need much time to think to follow the drone.\\
            \hline
            Physically easy to use& I don't need to put in a lot of physical effort when using the system.\\
            \hline
        \end{tabular}
        \large
        \label{tab:questionnaires}
    \end{table}
    \renewcommand{\arraystretch}{1}
    
\begin{figure}[!t]
        \centering
        \includegraphics[width=0.8\columnwidth]{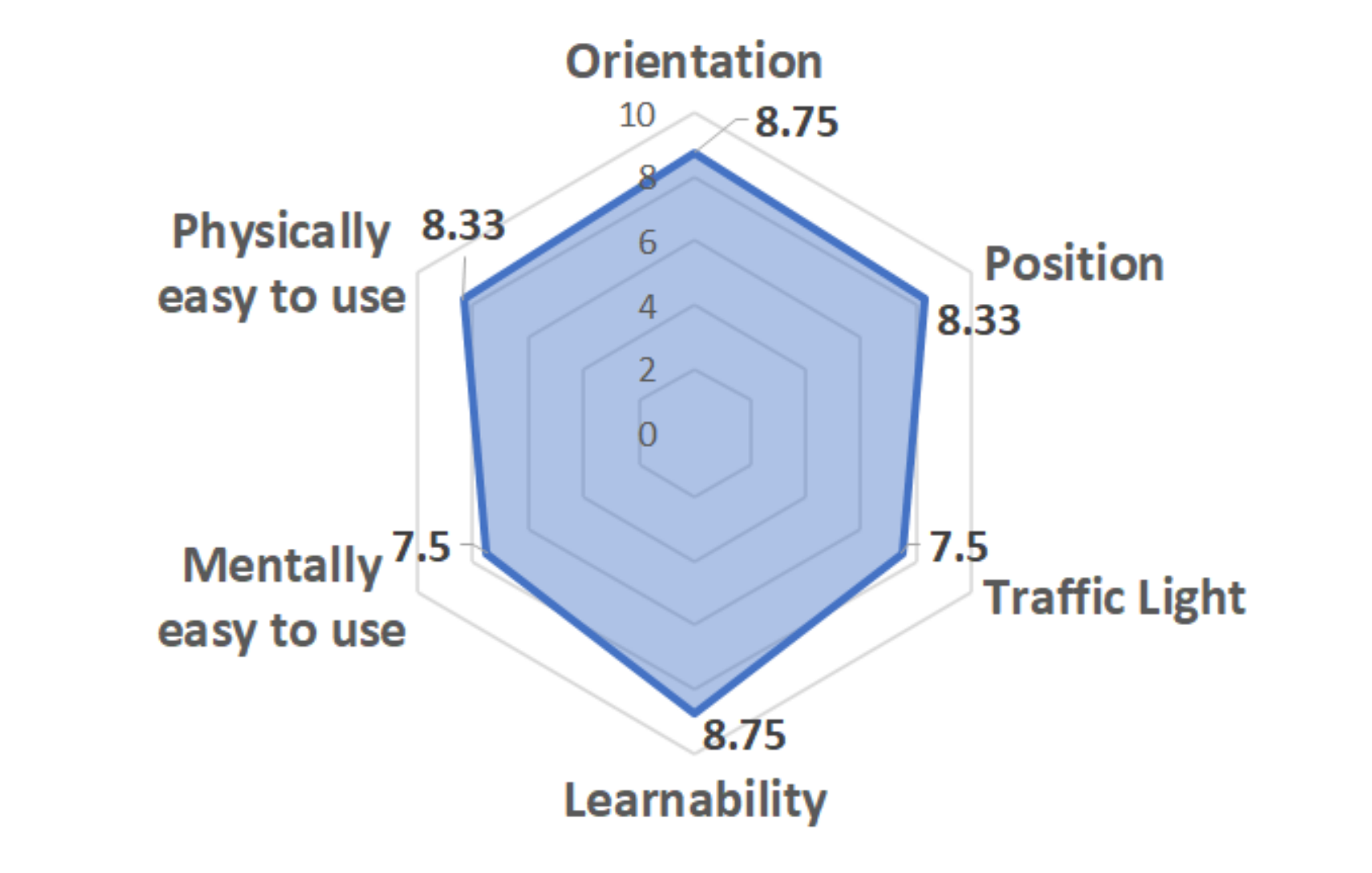}
        \vskip -2ex
        \caption{\small Evaluation results of the user study. Average scores are given by participants}
        \label{fig:user_study_scores}
    \end{figure}

    \noindent \textbf{Evaluation}. A radar graph assessing multiple aspects is shown in Fig. \ref{fig:user_study_scores}. For features of walking along the walkable path, \emph{Orientation} and \emph{Position}, participants find it reliable for guiding, citing that they can even avoid obstacles using this system. In terms of \emph{Physically easy to use} and \emph{Learnability}, they also make positive comments. After a brief introduction the participants are able to use the system alone. 
    However, the \emph{Traffic light} feature still needs to be improved. It could be harder to cross the street with a series of pedestrian lights for wrongly guiding users to the pedestrian zone in the middle. To tackle this issue, the speed of drone can be adjusted during crossing via the aforementioned drone control algorithm.
    The participants also rate the feature \emph{Mentally easy to use} lower than others, as sometimes it is insufficient to feel the traction from the connected string. This issue is subsequently improved with additional voice feedback using a wireless bone conduction headphone, which aims to alert users the walking direction and the color of the pedestrian light. 
    Despite a limited number of participants, most of them comment that the function of our prototype is indeed very helpful for people with visual impairment, which proves the usability and effectiveness of our flying guide dog prototype.
    
    \section{Conclusion}
\label{sec:conclusion}

    % Summary
    In this work, we propose a novel ``flying guide dog" prototype. Combining street view semantic segmentation, traffic light classification, and drone control algorithm, our prototype is capable to automatically discover the walkable path, avoid obstacles, and thus guide the user walk safely. Moreover, to better distinguish pedestrian traffic lights from other types of traffic lights, we set up a new dataset called \textit{\acrshort{pvtl}}. Results of the user study suggest that our approach is effective for \acrshort{bvip} assistance.
    
    % Limitations and future work
    Nonetheless, the prototype is limited by the drone's battery capacity, which can only support maximal 13 minutes flight time. Another major limitation is that the drone is too light to resist the wind. A reasonable approach to tackle these issues could be to use a more powerful drone with larger battery capacity. In the future, we plan to explore the usage of embedded AI computers such as Nvidia Jetson AGX Xavier and Jetson Nano to improve portability.

    \bibliographystyle{IEEEtran}
    \bibliography{literature}
    
\end{document}